# AITA Generating Moral Judgements of the Crowd with Reasoning

Ameer Sabri

Osama Bsher

## 1 Introduction

Morality is a fundamental aspect of human behavior and ethics, influencing how we interact with each other and the world around us. When faced with a moral dilemma, a person's ability to make clear moral judgments can be clouded. Due to many factors such as personal biases, emotions and situational factors people can find it difficult to decide their best course of action.

The AmITheA******[1] (AITA) subreddit is a forum on the social media platform Reddit that helps people get clarity and objectivity on their predicaments. In the forum people post anecdotes about moral dilemmas they are facing in their lives, seeking validation for their actions or advice on how to navigate the situation from the community. The morality of the actions in each post is classified based on the collective opinion of the community into mainly two labels, "Not The A******" (NTA) and "You Are The A******" (YTA).

This project aims to generate comments with moral reasoning for stories with moral dilemmas using the AITA subreddit as a dataset. While past literature has explored the classification of posts into labels (Alhassan et al., 2022), the generation of comments remains a novel and challenging task. It involves understanding the complex social and ethical considerations in each situation. To address this challenge, we will leverage the vast amount of data on the forum with the goal of generating coherent comments that align with the norms and values of the AITA community. In this endeavor, we aim to evaluate state-of-the-art seq2seq text generation models for their ability to make moral judgments similarly to humans, ultimately producing concise comments providing clear moral stances and advice for the poster.

## 2 Related Work

Research in the field of NLP has explored the concept of morality from various angles. However, much of the work focuses on different tasks than what our project aims to accomplish. For example, some studies, such as Fulgoni et al. (2016), use moral foundation measurement to score text based on five sets of moral intuitions (Graham et al., 2009). Other research has examined whether text expresses an opinion that is in favor of, against, or neutral towards a particular topic or idea (Mohammad et al., 2016).

Our project, on the other hand, focuses on analyzing the morality of individuals' actions in everyday situations, aiming to determine if their behavior is morally right or wrong. Previous research has mainly explored morality judgment in classification tasks, like predicting court decisions (Aletras et al., 2016), or within the AITA subreddit by classifying posts or comments as YTA or NTA (Alhassan et al., 2022; Botzer et al., 2021). Our project is distinct as it seeks to generate morality-related comments, which remains largely unexplored. The current state of the art for generating comments on AITA posts is ChatGPT-3[2]; however, it wasn't specifically trained for this task, and its underlying models are not publicly available. Consequently, we will investigate other state-of-the-art language models.

Recent advances in text generation have shown promising results. Kale and Rastogi (2020) fine-tuned T5 (Raffel et al., 2019) models for data-to-text generation, achieving state-of-the-art performance on task-oriented dialogue, tables-to-text, and graph-to-text tasks. Yermakov et al. (2021) conducts research in the medical domain, fine-tuning BART (Lewis et al., 2020) and T5 to generate multi-sentence texts given an input of medical entities. A more similar research paper uses GPT-2 (Radford et al., 2019) to generate tweets in the style of certain politicians (Ressmeyer, 2019). They train GPT-2 by prepending the context tweet (the tweet to which they want to generate a reply) to the actual reply, and include the tweeter's name in the input sample to generate tweets that mimic the styles of specific political figures.

To the best of our knowledge, these methods of transfer learning and fine-tuning for text-to-text generation have not been applied to the field of morality. We propose two contributions: collecting a new dataset of posts and comments from AITA that includes high-quality samples of a post and its top comments together with the verdict, and conducting research to generate morality-related comments using state-of-the-art transformers in a previously unexplored field.

---
[1] www.reddit.com/r/AmItheAsshole/

[2] chat.openai.com

## 3 Methodology

### 3.1 Dataset Collection

Given the absence of an existing dataset for the r/AITA subreddit containing both posts and comments, we created a novel dataset by following the approaches of Alhassan et al. (2022) and O'Brien (2020). While Alhassan et al. (2022) collected a dataset of 175,000 posts for a binary classification task, our goal was to create a more comprehensive dataset that included not only the posts but also their top comments.

After scraping the entire subreddit and applying pre-processing techniques, we retained only those posts with a certain score threshold, resulting in a final dataset comprising 270,709 entries. Although our primary focus is on text generation, this dataset can also be leveraged for text classification tasks, similar to the work done by Alhassan et al. (2022). Given its larger size and coverage of the entire subreddit, we anticipate that our dataset could potentially yield improved results across various tasks.

In this section, we provide a detailed overview of the multiple steps involved in collecting and curating the r/AITA dataset. The first step in our workflow as illustrated in Figure 2 in the appendix.

#### 3.1.1 Pushshift API

To collect the post IDs from the AITA subreddit, we utilized the Pushshift API[3], a widely adopted tool for accessing Reddit data. This API allowed us to gather all post IDs from the inception of the subreddit up until April 1st, 2023. One of the primary advantages of using the Pushshift API over the official Reddit API (PRAW) is its ability to efficiently retrieve posts within a specific time range. The official Reddit API does not provide an easy way to query posts based on their creation timestamps. In contrast, Pushshift enables researchers to specify a time range using "after" and "before" parameters, allowing for a more targeted and efficient data collection process.

---
[3]https://github.com/pushshift/api

Our script was able to collect 1.6 million post IDs. After obtaining the post IDs from the Pushshift API, we then employ PRAW to fetch the actual content of the posts.

#### 3.1.2 PRAW

Utilising the list of IDs collected with Pushshift, PRAW[4] was used to fetch the post title, body, verdict, comments, as well as other post entities. An example of the relevant post entities for training can be found in Table 3 in the appendix. To fetch all 1.6 million posts, one hundred bots were created to scrape the data in parallel; this is because PRAW limits the number of requests that can be sent per minute, accelerating the procedure from 33 days to eight hours. Posts that do not contain a verdict and had fewer than 2 comments were skipped, as they cannot be used for training. Afterward, the posts collected by all the bots were merged into a single file containing 817,661 posts. To prepare the data for training, data cleaning and pre-processing techniques were employed, which are explained in the following section.

#### 3.1.3 Data Cleaning and Pre-processing

Data exploration was carried out to determine the most effective cleaning and pre-processing methods for the dataset. As a result of this exploration, we removed empty, deleted, or posts without a verdict. We also eliminated new line characters, post edits and updates, and converted all letters to lowercase. To avoid training the model to generate brief responses, posts with fewer than ten tokens and comments with fewer than five tokens were removed. Additionally, we filtered out posts where the verdict differed from the top two comments to ensure that the comments aligned with the assigned post verdict.

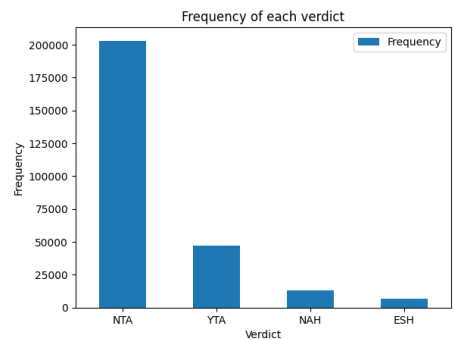

**Figure 1:** The figure illustrates the frequency of each verdict in the dataset

The dataset exhibited an imbalanced distribution of post verdicts, as shown in Figure 1. To enable the model to generate comments for various verdicts found in the subreddit, we combined NTA verdicts with NAH (No A\*\*holes Here) verdicts, and YTA verdicts with ESH (Everyone Sucks Here) verdicts, as illustrated in Table 1. Subsequently, we divided the dataset into eight balanced subsets for experimentation using label "0" (NTA and NAH) undersampling to mitigate majority class bias, as depicted in Table 4 in the Appendix.

---
[4]https://praw.readthedocs.io/en/stable/

| Verdict | Frequency | Label | Frequency |
|---|---|---|---|
| NTA (Not The A****) | 203,079 | 0 | 216,421 |
| NAH (No A**** here) | 13,342 | | |
| YTA (You Are The A****) | 47,408 | 1 | 54,288 |
| ESH (Everyone Sucks here) | 6,880 | | |

Table 1: The table illustrates the mapping frequency of each verdict and the label encompassing the verdicts

Considering the differences in maximum input sequence length between T5 and BART models (Raffel et al., 2019; Lewis et al., 2020), the data was further separated based on post lengths. Additionally, we split the dataset into posts with and without titles to examine their potential impact on model performance. Lastly, two subsets were created: one containing a single comment per post and another with two comments per post. This allowed us to evaluate the effect of a larger (double) dataset on model performance.

### 3.2 Metrics

#### 3.2.1 Accuracy

In typical text generation tasks, accuracy is not a standard metric. However, for our specific case, we have devised an accuracy measure that evaluates whether the model successfully matches the overall verdict of the generated comment with the actual comment. To do this, we focus on whether the generated comment correctly reflects the YTA (You're The A\*\*hole) or NTA (Not The A\*\*hole) verdict of the original comment.

Our approach involves extracting the YTA/NTA portion of the generated comment and comparing it with the label of the actual comment. This allows us to assess the model's ability to generate contextually accurate comments that align with the sentiment and judgment of the original content.

#### 3.2.2 Automated Evaluation Metrics

BERTScore (Zhang et al., 2019) is an evaluation metric that utilizes the contextual embeddings of BERT (Devlin et al., 2018) to measure the similarity between generated and reference texts. By comparing token embeddings, BERTScore captures both semantic and syntactic information, making it more effective than traditional n-gram based metrics like BLEU. BERTScore has demonstrated a better correlation with human judgments across various NLP tasks.

BARTScore (Yuan et al., 2021), similar to BERTScore, employs the embeddings of the BART model (Lewis et al., 2020) instead of BERT. BARTScore benefits from BART's excellent performance in text generation tasks, as it is specifically designed for sequence-to-sequence problems. This makes BARTScore more suitable for evaluating generated text in tasks like text summarisation, translation, and comment generation.

BLEU (Papineni et al., 2002) measures the similarity between generated and reference texts based on n-gram overlap and is primarily used for machine translation tasks. METEOR (Banerjee and Lavie, 2005) addresses some of BLEU's limitations by incorporating synonym matching and sentence structure, resulting in improved correlations with human judgments. ROUGE (Lin, 2004), typically applied in text summarisation tasks, measures similarity by comparing n-grams, word sequences, and word pairs. These traditional automated metrics have their own strengths and weaknesses.

It's important to note that traditional metrics like BLEU, METEOR, and ROUGE have limitations in providing a comprehensive analysis of generated comments, particularly because they are primarily designed for tasks such as text summarisation and translation. We anticipate that BERTScore and BARTScore, which are specifically tailored to capture semantic and syntactic information, will outperform these traditional metrics for our task. Nevertheless, we compute all these metrics to assess their informativeness in evaluating generated comments within the context of r/AITA posts.

#### 3.2.3 Human Evaluation

To gain a more comprehensive understanding of our model's generated outputs, we conducted a survey to compare the generated comments against actual human-written comments. In the survey, participants were asked to read a post from r/AITA and evaluate three associated comments. These comments included one human-written comment and one comment each from our best T5 and BART models. Participants were then asked to rank the comments from 1 to 3 and identify the one they believed to be human-generated. The survey featured five posts in total, comprising three NTA and two YTA posts. To encourage participation, we kept the survey relatively short. The link to the survey is provided in the footnote[5].

## 4 Experiments

In recent years, Transformer models based on Google's architecture (Vaswani et al., 2017) have become the standard for state-of-the-art performance in natural language processing tasks. Text generation, in particular, has gained widespread attention with the release of models like ChatGPT. In our experiments, we explore the application of two cutting-edge text generation models, BART and T5, to the unique task of generating comments for posts on the r/AITA subreddit. To the best of our knowledge, this specific task has not been previously attempted using either of these models, making our research novel.

### 4.1 Experiment Settings

We utilised the Hugging Face[6] implementations of BART and FLAN-T5 for our project. Our code was implemented using PyTorch[7], leveraging Hugging Face's utility functions for training, tokenization, and evaluation. We appreciate the support from Research IT and the University of Manchester's Computational Shared Facility, whose provision of two NVIDIA V100 GPUs was essential for our computationally intensive text generation project.

We employed a learning rate of 2e-5 using the AdamW optimizer and set the batch size to 2 to avoid out-of-memory errors. The maximum sequence length was set to their respective limits, 512 for T5 and 1024 for BART. We configured the training to run for 30 epochs and used early stopping to mitigate overfitting. The cross-entropy loss function, which is applied over the generated token probabilities and target token ids, is optimized during training to minimize the difference between the generated sequence and the target sequence.

We utilized Optuna (Akiba et al., 2019) for automated hyperparameter optimization, focusing solely on the learning rate due to memory constraints that prevented increasing the batch size. Early stopping was employed, rendering the number of epochs a non-variable. However, we quickly discovered that running 30 trials for BART on the smallest dataset would take approximately 300 hours, or 12 days. Extrapolating this to each model and dataset, the largest T5 dataset would require a staggering 87 days. Consequently, we opted for a fixed learning rate in our experiments considering the time constraints.

| Model | Total Parameters | Corpus Size |
|---|---|---|
| T5 | 250 million | 750GB |
| BART | 140 million | 160GB |

Table 2: Total Parameters of pre-trained language models

For dataset preparation, we employed Scikit-learn's train-test split[8] function to randomly divide our datasets into training, testing, and validation sets. We allocated 80% of each dataset for training and 10% each for testing and validation. Table 2 provides an overview of the models, including their number of parameters and the corpus size on which they were trained. Due to GPU constraints, we opted for the base versions of the models instead of their larger counterparts.

### 4.2 Finetuning T5

In our experiments, we employ the T5 (Text-to-Text Transfer Transformer) model, specifically the FLAN-T5 Base variant, for fine-tuning on our dataset. The original T5 model was released in 2019 by Raffel et al. (2019). It is a powerful and versatile model based on the Transformer architecture that is effective on a variety of NLP tasks and can be down streamed for custom tasks. The primary goal of using the FLAN technique is to reduce the time required for adaptation to new languages or new domains, making the model more efficient when fine-tuning on specific tasks or datasets (Chung et al., 2022). We opted to use the FLAN version of the model as it has proved to perform better than the original in multiple tasks (Chung et al., 2022). The FLAN-T5 can handle a maximum input of 512 tokens. We made use of the huggingface **T5ForConditionalGeneration** class for finetuning the model to our task.

We hypothesise that T5, having more model parameters and a larger pre-training corpus, should perform better on this task than BART, especially when working with datasets containing shorter posts. This is because T5 has a limitation in handling token sequences longer than 512. We also expect that the model will show better performance on datasets with double comments, as it can benefit from the extra context provided by the additional comment. Moreover, we believe that including the post title in the input will give the model useful information about the post, allowing it to generate more relevant and context-aware comments. Overall, our hypothesis suggests that T5 will be effective in generating comments for r/AITA posts when given enough context and when working with shorter posts.

---

[5]https://forms.gle/zx7ShNyNDFSCaHXS9
[6]https://huggingface.co/
[7]https://pytorch.org/
[8]https://scikit-learn.org

### 4.3 Finetuning BART

BART (Lewis et al., 2020), a powerful pre-trained model introduced by Facebook in 2019, leverages bidirectional context during pretraining and learns by reconstructing original sentences from corrupted versions using denoising autoencoding. This approach enables BART to develop stronger text representations. BART's maximum sequence length of 1024 input tokens allows it to handle longer input sequences compared to T5. We use the **BARTForConditionalGeneration** class from Hugging Face to fine-tune the model for our task.

Given BART's ability to handle longer sequence lengths, we expect it to outperform T5 on datasets with longer posts. We also anticipate that incorporating post titles and using double comments in the dataset will improve results. By doubling the dataset size with double comments, BART may gain more context and training examples. As a result, we hypothesise that BART will demonstrate superior performance compared to T5, particularly when dealing with extended text input.

## 5 Results and Discussion

Table 5 in the appendix presents the results of training our models on various datasets. It's worth noting that the BLEU Score is consistently 0 for all models. BLEU Score, which relies on n-gram overlap between generated and actual comments, appears inadequate for our task, as our models generate semantically similar comments without using the exact words. This is consistent with BLEU Score's primary design for machine translation tasks. Meteor Score, while slightly higher than BLEU, also utilizes a similar matching scheme based on exact words but includes synonym matching. This reinforces our belief that the model produces semantically related comments with few shared n-grams. Additionally, the discrepancy between actual comment lengths (average of 50) and those generated by the Trainer function (max length of 20) during training could contribute to BLEU and Meteor's limitations for our task.

BERTScore and BARTScore, which prioritize capturing meaning between generated and actual comments, are more suitable metrics. Length differences are less impactful, provided the generated comment conveys the same meaning as the label. Both T5 and BART exhibit similar BERT and BART Scores, indicating their proficiency at capturing meaning. In contrast, the accuracy metric is more informative for model comparison, with BART significantly outperforming T5, although T5 required at least triple the training time across all the subsets. Our best model, BART, achieved **85%** accuracy and a BERTScore of **0.75** on the dataset with double comments and longer posts. The accuracy of BART rivals that of the current best, 88%, in the moral judgement classification for the r/AITA subreddit (Alhassan et al., 2022). This aligns with our hypothesis that longer posts provide more context, enabling the model to learn and generalize better. However, the highest accuracy occurred in the dataset without titles, contradicting our initial hypothesis. Overall including or excluding the title from the post did not consistently influence performance, however, the subsets with longer posts and double the number of comments per post had better metrics consistently, this could be due to the larger size of these subsets as illustrated in Table 4 in the appendix.

BART consistently outperforms T5, which could be attributed to its pretraining method and architecture. BART's denoising autoencoder objective enables the model to learn more meaningful text representations, and its bidirectional encoder facilitates context learning from both directions (Lewis et al., 2020). Nonetheless, we believe that automated metrics alone are insufficient to conclusively determine whether T5 or BART is the superior model for our task (Callison-Burch et al., 2007). To complement our quantitative analysis, we conducted human evaluations. Studies have shown that using both human evaluation and automated metrics is the most effective approach for assessing text-generation tasks (Belz and Reiter, 2006).

In total, we received 47 responses in one week for the human survey, which enabled us to assess whether the comments generated by our models were on par with human-generated comments and if they could pass as such. The respondents included members of the r/SampleSize subreddit[9], a community dedicated to participating in surveys, as well as fellow students and

---
[9] https://www.reddit.com/r/SampleSize/

NLP experts.

On average, participants were unable to differentiate between human responses and those generated by T5 and BART models. When asked to identify the human comment across five posts, the average participant correctly identified 2 human comments. No participant managed to correctly identify all human comments across the posts. These findings are presented in Table 3 in the appendix.

Additionally, when asked to rank the comments for each post, participants rated the human response as the best for three out of the five posts. BART-generated comments were ranked highest for two out of the five posts, while T5-generated comments were not ranked first for any of the posts. The comment rankings are illustrated in Figure 4. The overall percentage of times a human comment was selected as best across the five posts was approximately 43.6%, while BART-generated comments were selected 31.9% of the time, and T5-generated comments were chosen as the best 24.3% of the time, as shown in Figure 5. The discrepancy between the best comment and human comment selection could be influenced by the perception of AI-generated responses from usage of GPT-based language models. Some respondents might be familiar with the capabilities of GPT-based models and attribute formal and neutral comments to AI-generated comments. Nevertheless, due to being trained on the AITA subreddit, the comments generated by BART and T5 are blunt and less neutral in nature, which could have an impact on the evaluation results.

The results indicate that participants found human comments to be the most appropriate overall, followed closely by BART-generated comments, with T5-generated comments being the least favoured. Similar to other evaluation metrics, the human survey suggests that BART outperforms T5. Overall the results suggest that model-generated comments can pass as human comments and can provide a better response than human responses for posts in the subreddit as illustrated in Figure 6.

## 6 Conclusion and Future Work

In this study, we have tackled the challenge of generating comments with moral reasoning for stories with moral dilemmas using the AITA subreddit as a dataset. By employing state-of-the-art transformer models, BART and T5, we have demonstrated the potential for generating coherent comments that align with the norms and values of the AITA community. Our findings suggest that BART outperforms T5 in comment generation for posts in the AITA subreddit, as indicated by quantitative and qualitative metrics, and is also less computationally expensive to train and evaluate.

Our contributions in this project include collecting a novel dataset of posts and comments from AITA, including high-quality samples of a post and its top comments together with the verdict, and conducting research to generate morality-related comments using state-of-the-art transformers in a previously unexplored field.

Future work should focus on testing other state-of-the-art models, such as GPT-2 and BLOOM, and conducting hyperparameter tuning for both BART and T5 with larger model variations. Additionally, creating a dataset consisting primarily of longer comments might help train the models to be better at generating more detailed and extensive comments, which could be an interesting future experiment. Further investigation should be conducted to find more appropriate evaluation metrics for assessing the generated comments, either through other metrics or by altering the model training process. Increasing the number of respondents and ensuring a greater diversity of participants in human surveys would also help improve the confidence in the evaluation of the generated comments.

Another important goal for future work includes publishing both the dataset collected and the models trained in this report, allowing for peer review of the findings, comparison of different models on the same dataset, and exploration of different evaluation metrics.

# 7   Appendix

| Header  | Example                                                                 |
|---------|-------------------------------------------------------------------------|
| Title   | AITA for playing video games while wife complains                        |
| Body    | Let me start by saying I'm not one who normally sits... AITA for wanting to play games? |
| Verdict | A**hole                                                                  |
| Comment | YTA. Video games come after the stuff you're supposed to do...           |

Table 3: The table illustrates an example post structure, with it's title, body, verdict and an example comment

| Dataset | Word count | Comments per post | Verdict Size |
|---------|------------|-------------------|--------------|
| With Title | >10 words <512 words | Single | 0: 39,939 <br> 1: 39,939 |
| | | Double | 0: 79,878 <br> 1: 79,878 |
| | >10 words <1024 words | Single | 0: 53,976 <br> 1: 53,976 |
| | | Double | 0: 107,952 <br> 1: 107,952 |
| Without Title | >10 words <512 words | Single | 0: 40,950 <br> 1: 40,950 |
| | | Double | 0: 81,900 <br> 1: 81,900 |
| | >10 words <1024 words | Single | 0: 53,992 <br> 1: 53,992 |
| | | Double | 0: 107,984 <br> 1: 107,984 |

Table 4: The table illustrates the different datasets and their balanced sizes

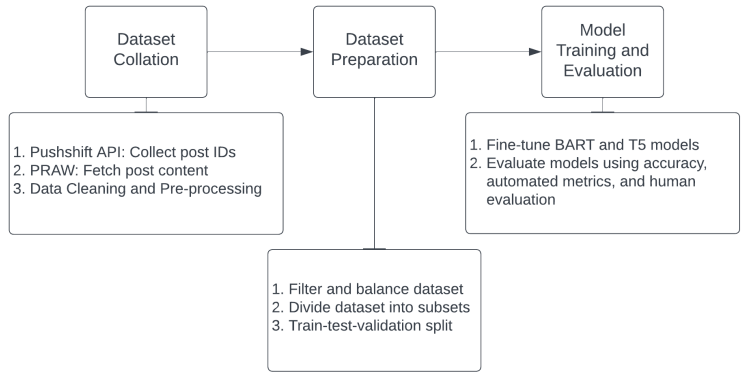

**Figure 2:** The flow chart depicts the overall methodology

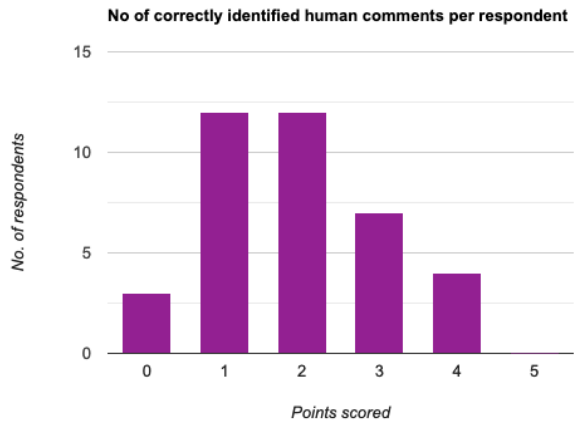

**Figure 3:** The bar chart illustrates the number of correctly identified human comments per respondent out of five

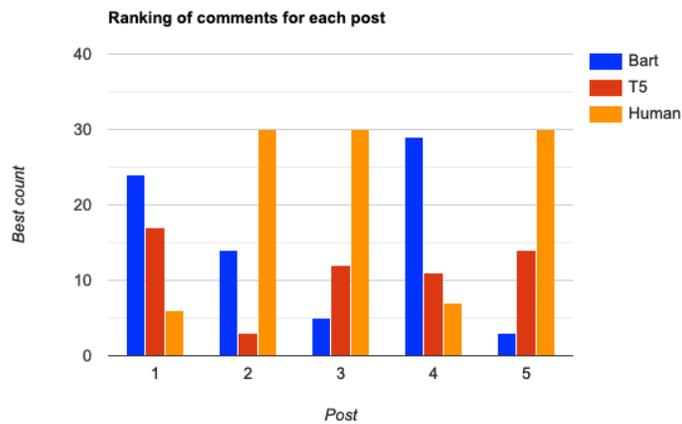

**Figure 4:** The bar chart illustrates the frequency the comment is selected as the best response for each post

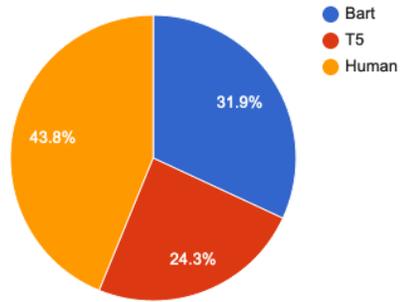

**Figure 5:** The pie chart illustrates the percentage of time each comment type was selected as the best comment for all posts

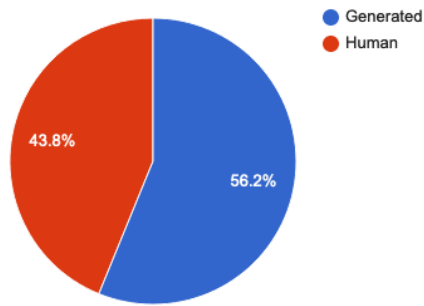

**Figure 6:** The pie chart illustrates the percentage of time each comment type was selected as the best comment for all posts, humans against generated comments

| Dataset | Word count | Comments per post | Model | Accuracy | BLEU Score | Meteor Score | Bert Score | Bart Score | Rogue 1 Score | Rogue 2 Score | Epochs | Training Time (h) |
|---|---|---|---|---|---|---|---|---|---|---|---|---|
| With Title | >10 words <512 words | Single | T5 | 0.74 | 0.0 | 0.08 | 0.74 | -6.64 | 0.14 | 0.02 | 13 | 40.7 |
| | | | Bart | 0.75 | 0.0 | 0.09 | 0.75 | -6.57 | 0.17 | 0.03 | 6 | 9.4 |
| | | Double | T5 | 0.79 | 0.0 | 0.08 | 0.75 | -6.64 | 0.15 | 0.02 | 13 | 72.2 |
| | | | Bart | 0.81 | 0.0 | 0.1 | 0.75 | -6.55 | 0.17 | 0.03 | 6 | 19.9 |
| | >10 words <1024 words | Single | T5 | - | - | - | - | - | - | - | - | - |
| | | | Bart | 0.77 | 0.0 | 0.09 | 0.75 | -6.63 | 0.16 | 0.03 | 6 | 20.7 |
| | | Double | T5 | - | - | - | - | - | - | - | - | - |
| | | | Bart | 0.83 | 0.0 | 0.09 | 0.75 | -6.66 | 0.16 | 0.03 | 6 | 35.0 |
| Without Title | >10 words <512 words | Single | T5 | 0.73 | 0.0 | 0.08 | 0.74 | -6.68 | 0.14 | 0.02 | 13 | 43.9 |
| | | | Bart | 0.73 | 0.0 | 0.09 | 0.75 | -6.60 | 0.16 | 0.02 | 6 | 10.1 |
| | | Double | T5 | 0.79 | 0.0 | 0.08 | 0.74 | -6.64 | 0.14 | 0.02 | 13 | 77.2 |
| | | | Bart | 0.79 | 0.0 | 0.09 | 0.75 | -6.58 | 0.16 | 0.03 | 6 | 17.5 |
| | >10 words <1024 words | Single | T5 | - | - | - | - | - | - | - | - | - |
| | | | Bart | 0.75 | 0.0 | 0.09 | 0.75 | -6.67 | 0.15 | 0.02 | 6 | 22.2 |
| | | Double | T5 | - | - | - | - | - | - | - | - | - |
| | | | Bart | **0.85** | 0.0 | 0.09 | 0.75 | -6.63 | 0.16 | 0.03 | 8 | 42.1 |

Table 5: The table illustrates the results for all the experiments.